\documentclass{article}

\PassOptionsToPackage{numbers, compress}{natbib}
 \usepackage[preprint]{neurips_2026}

\usepackage[utf8]{inputenc} 
\usepackage[T1]{fontenc}    
\usepackage{hyperref}       
\usepackage{url}            
\usepackage{booktabs}       
\usepackage{amsfonts}       
\usepackage{nicefrac}       
\usepackage{microtype}      
\usepackage{xcolor}         
\usepackage{amsmath}
\usepackage{graphicx}
\usepackage{subcaption}
\usepackage{algorithm}
\usepackage{listings}
\usepackage{wrapfig}
\usepackage{multirow}

\title{Beyond Point-Wise Matching: Structural Representation Alignment for Accelerating Diffusion Transformers}

\author{
\textbf{Shaodong Xu}$^{1}$\thanks{Equal contribution.} \quad
\textbf{Zhendong Wang}$^{1}$\footnotemark[1] \quad
\textbf{Litong Gong}  \quad
\textbf{Zexian Li} \quad
\textbf{Wengang Zhou}$^{1}$ \\
\textbf{Tiezheng Ge} \quad
\textbf{Houqiang Li}$^{1}$ \\
$^{\text{1}}$ University of Science and Technology of China \quad
}

\begin{document}

\maketitle

\vspace{-9pt}
\begin{abstract}
Recent advances in Diffusion Transformers (DiTs) demonstrate that aligning noisy latent states with well-trained semantic features—as pioneered by Representation Alignment (REPA)—can substantially accelerate training and improve generation fidelity.
Subsequent analysis (\textit{e.g.}, iREPA) suggests that these gains arise primarily from transferring spatial structure contained in pre-trained vision representations. However, mostly existing alignment methods employ point-wise matching objectives or rely on implicit architectural tweaks, which fail to explicitly model the spatial relational geometry inherent in vision foundation models.
We argue that such element-wise supervision is insufficient to capture the rich spatial topology of visual representations, and that effective alignment for generation should instead be formulated as an explicit structural constraint.
To this end, we propose sREPA, a \textbf{s}tructural \textbf{REP}resentation \textbf{A}lignment framework to enforce consistency in the relational geometry of feature maps, rather than merely matching individual feature points.
By encouraging the model to internalize holistic spatial layouts and structural correlations from pre-trained features, sREPA achieves faster and more stable convergence, along with improved sample quality, compared to state-of-the-art alignment strategies.
Our code and models will be released.
\end{abstract}

\section{Introduction}
\label{sec:intro}
Denoising generative models~\citep{ddpm, score,flowmatching,kobyzev2020normalizing} have become the dominant paradigm in high-fidelity visual synthesis~\citep{sdxl, sd3, flux2023, brooks2024video,wu2025qwen}. Among them, scaling diffusion transformers has demonstrated remarkable capacity and scalability~\citep{dit,sit}. However, training these models from scratch remains computationally expensive. 
A central challenge is that the standard denoising objective provides limited direct supervision for learning meaningful visual representations, which may benefit the generation quality.

\begin{figure*}[t]
\centering
\includegraphics[width=\linewidth]{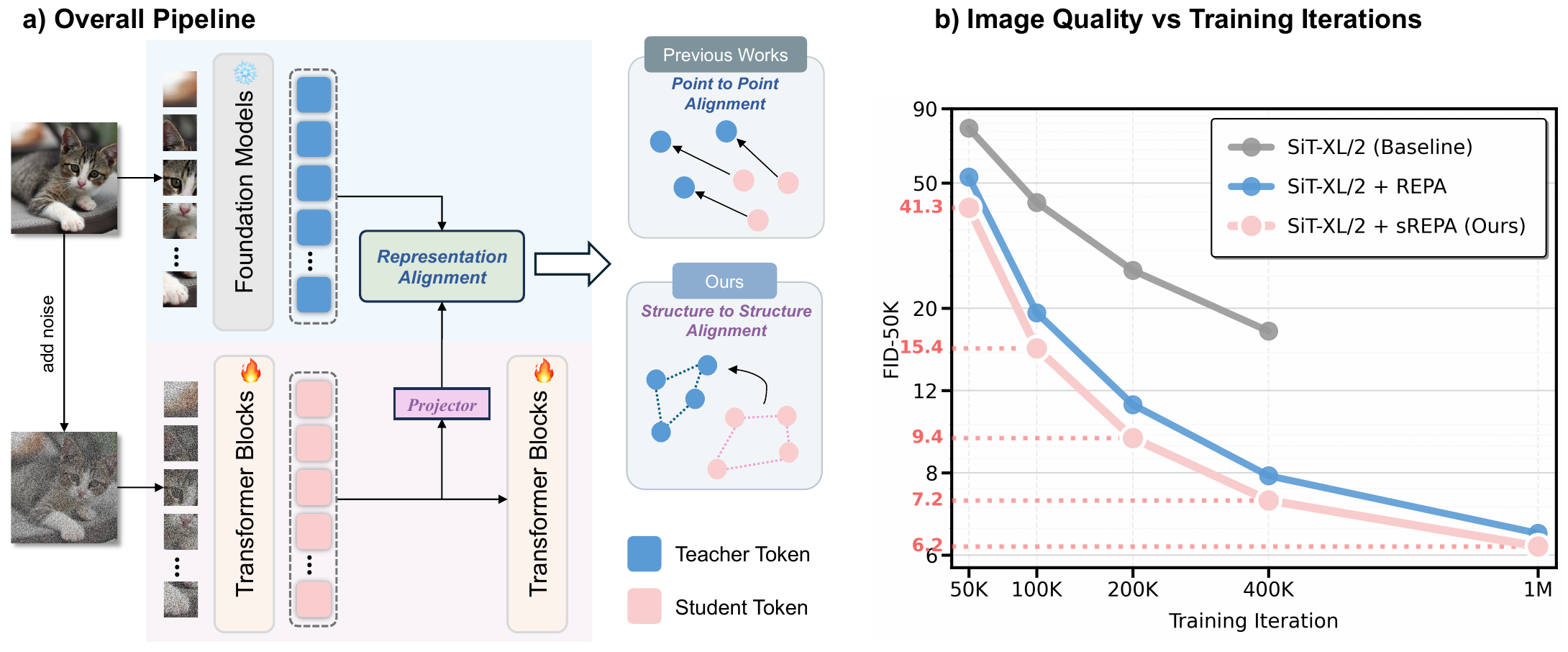}
\caption{
\textbf{Structural representation alignment improves diffusion model training beyond point-wise alignment.}
Compared with the standard REPA training framework, the proposed \textbf{sREPA}, integrates explicit structural supervision to align relational distributions between teacher and student representations, resulting in faster convergence and consistently better generation quality.}
\label{fig:teaser}
\vspace{-9pt}
\end{figure*}

The pioneer work, representation alignment~(REPA) is emerged as a powerful technique for accelerating the training of diffusion transformers~\citep{repa}. By aligning diffusion internal representations with features extracted from pre-trained self-supervised visual encoders, REPA-style methods achieve improvements in convergence speed and final generation performance~\citep{repa-e,reg,kouzelis2025boosting}. Despite their empirical gains, most existing alignment strategies primarily rely on point-wise regression, which enforces local feature consistency by minimizing distances between individual corresponding tokens in the student and teacher feature maps. 

A recent study iREPA~\citep{irepa} provides further insight into the mechanism behind representation alignment, showing that its effectiveness arises primarily from the transfer of spatial structure rather than high-level semantic content. To support this observation, they introduce architectural modifications, including a convolutional projector and spatial normalization, to better preserve spatial inductive biases when aligning diffusion representations with pre-trained visual features. Their results demonstrate that enhancing the accessibility of spatial structure indeed leads to improved generative performance.

However, despite emphasizing the importance of spatial structure, iREPA still formulates alignment using conventional point-wise matching objectives, where supervision is applied independently to individual feature tokens. Such element-wise alignment fails to explicitly model the relational geometry and global topology encoded in foundation model representations, leaving the structural consistency between feature maps only implicitly enforced.

In this paper, we propose sREPA (\textbf{s}tructural \textbf{REP}resentation \textbf{A}lignment), a simple yet effective framework that introduces explicit structural supervision into representation alignment. As illustrated in Figure~\ref{fig:teaser}, unlike prior approaches that emphasize isolated token-level matching, sREPA fundamentally shifts the supervision granularity from individual feature points to their pairwise relational structure. Our core insight is that \textit{for a generative model to fully exploit the structural priors encoded in a pre-trained vision backbone, it is insufficient to merely align corresponding feature tokens}. Instead, the model must internalize the encoder’s underlying \emph{structural geometry}, \textit{i.e.}, the relative relationships and global topology that govern how different image regions are organized in the representation space. 

To this end, the proposed sREPA aligns the student’s internal similarity structure with that of the teacher by matching their similarity distributions, thereby explicitly preserving the relational topology of latent feature maps. We instantiate this principle through two practical variants: \emph{sREPA-MSE}, a parameter-free formulation that directly regresses pairwise relations, and \emph{sREPA-KL}, a probabilistic variant that performs distribution-level structural matching.

Extensive experimental results with detailed analysis demonstrate that sREPA consistently improves generation quality while significantly accelerating convergence, outperforming existing point-wise alignment strategies.

Our contributions can be summarized as follows:
\begin{itemize}
    \item We identify a fundamental limitation of existing representation alignment methods, showing that commonly used point-wise objectives are insufficient for explicitly capturing the spatial relational structure encoded in pre-trained visual encoders.
    \item We propose sREPA, a simple yet effective structural representation alignment framework that introduces structure-aware supervision, explicitly distilling relational topology from vision foundation models into diffusion transformers.
    \item Through extensive experiments, we demonstrate that explicit structural alignment substantially improves the utilization of spatial information, resulting in faster and more stable convergence as well as consistently superior generation quality.
\end{itemize}

\section{Related Work}
\label{sec:related}
\textbf{{Visual Generative Models.}} 
Diffusion Probabilistic Models (DPMs) synthesize high-fidelity images by iteratively denoising~\citep{ddpm,song2020denoising}. Early DPMs operate directly in pixel space, incurring high computational costs. Latent Diffusion Models (LDMs)~\citep{ldm} mitigate this by performing generation in the compressed latent space of a variational autoencoder. Subsequently, the backbone of diffusion models shifts from U-Nets\citep{ronneberger2015u} to transformers~\citep{dit}, while SiT~\citep{sit} further improved the efficiency by integrating interpolant-based Rectified Flow~\citep{flowmatching,flowstraight}. However, these transformer-based generators typically require massive training iterations to achieve convergence. To mitigate this challenge, masked modeling strategies~\citep{maskdit,gao2023mdtv2} are introduced to accelerate diffusion training. Despite their effectiveness, these methods often require modifications to the model architecture or masking strategies. In contrast, our sREPA possesses significantly faster convergence and superior generation performance while leaving the base model architecture unchanged, demonstrating that superior training dynamics can be reached without structural compromises.

\textbf{{Representation Learning for Visual Generation.}}
Recent advances in generative modeling have increasingly focused on leveraging the rich semantic priors of pre-trained Vision Foundation Models (VFMs). One popular research direction is aligning VAE. For example, VA-VAE~\citep{vavae} aligns the latent representations of VAE with pre-trained vision foundation models, significantly improving the final generation performance while preserving their original reconstruction capability. RAE~\citep{rae} moves further and directly replaces the original VAE encoder with a discriminative semantic encoder, forming high-dimensional semantically rich and structurally coherent latent spaces and resulting in a better generation capability, which also demonstrates the capability of self-supervised vision foundation models. Another promising paradigm to improve generation quality is to align the internal features of the generators with those of a powerful vision encoder. REPA~\citep{repa} pioneers this direction by aligning diffusion features with representations of a frozen teacher, demonstrating significant generation performance gains. A later work iREPA~\citep{irepa} improves REPA further by encouraging spatial consistency through convolutional projections and spatial normalization. DDT~\citep{wang2025ddt} decouples DiT into an encoder–decoder and applies REPA loss to the encoder. REG~\citep{reg} introduces a tunable class token into the DiT sequence and explicitly aligns it with external representations. Our work also belongs to this paradigm; however, our focus is on the analysis of the commonly-used point-wise alignment and further presents a novel structural alignment strategy.

\textbf{Knowledge Distillation and Structural Matching.}
Knowledge Distillation (KD) focuses on transferring knowledge from a teacher model to a student model~\cite{hinton2015distilling}. A line of work explores structural distillation strategies that preserve the relational geometry of teacher representations to improve knowledge transfer. Relational Knowledge Distillation (RKD)~\cite{park2019relational} introduces distance-wise and angle-wise objectives to transfer relational information, while similarity-preserving distillation~\cite{tung2019similarity} matches pairwise similarity matrices between teacher and student activations. Correlation Congruence for Knowledge Distillation (CCKD)~\cite{peng2019correlation} further matches teacher-student correlation matrices to transfer inter-instance relational knowledge. Despite these advances in knowledge distillation and relational structure matching, most existing structural alignment methods are developed for discriminative tasks. In contrast, our work focus on integrating explicit structural supervision into the training objectives for generative models, particularly diffusion transformers, to improve training dynamics and generation performance.


\section{Method}
\label{sec:method}
In this section, we first review the preliminaries of our work in Section \ref{subsec:preliminaries}; then analyze the limitations of existing alignment strategies in Section \ref{subsec:motivation}; finally, in Section \ref{subsec:srepa}, we present the formulation of sREPA and its two implementation variants.

\subsection{Preliminaries}
\label{subsec:preliminaries}
Our base model is mainly SiT~\citep{sit}, which is highly relevant to diffusion models and flow matching models, which share the same paradigm of transforming a simple prior distribution into a complex data distribution.
For simplicity, we mainly review the formula of flow matching.
Under the Conditional Flow Matching (CFM) formulation, the commonly-used flow process is a linear interpolation between a data sample $\mathbf{x}_1 \sim q(\mathbf{x})$ and a noise sample $\mathbf{x}_0 \sim \mathcal{N}(\mathbf{0}, \mathbf{I})$. Then the intermediate state $\mathbf{x}_t$ at time $t \in [0, 1]$ is defined as:
\begin{equation}
    \mathbf{x}_t = t \mathbf{x}_1 + (1 - t) \mathbf{x}_0,
    \label{eq:fm_interpolation}
\end{equation}
where $t=0$ and $1$ correspond to pure noise and clean data, respectively. 
The dynamics of this process are governed by an Ordinary Differential Equation (ODE):
\begin{equation}
    \frac{\mathrm{d}\mathbf{x}_t}{\mathrm{d}t} = \mathbf{v}_t(\mathbf{x}_t),
    \label{eq:ode}
\end{equation}
where $\mathbf{v}_t(\cdot)$ is a time-dependent vector field that flows the probability density. 

To learn the generative flow, a neural network $\mathbf{v}_\theta(\mathbf{x}_t, t)$ is trained to approximate the conditional vector field generating the path. Taking the time derivative of $\mathbf{x}_t$, the target velocity field is simply $\mathbf{u}_t(\mathbf{x} | \mathbf{x}_1) = \mathbf{x}_1 - \mathbf{x}_0$. The objective function is optimized by minimizing the mean squared error between the model prediction and this target velocity, formulated as:
\begin{equation}
    \mathcal{L}_{\text{FM}}(\theta) = \mathbb{E}_{t, \mathbf{x}_0, \mathbf{x}_1} \left[ \left\| \mathbf{v}_\theta(\mathbf{x}_t, t) - (\mathbf{x}_1 - \mathbf{x}_0) \right\|^2 \right],
    \label{eq:fm_loss}
\end{equation}
where $t \sim \mathcal{U}[0, 1]$. 
To generate samples, the trained model starts from $\mathbf{x}_0 \sim \mathcal{N}(\mathbf{0}, \mathbf{I})$ and numerically integrates Eq.~\eqref{eq:ode} from $t=0$ to $1$.

\begin{figure*}[t]
    \centering
    \includegraphics[width=\linewidth]{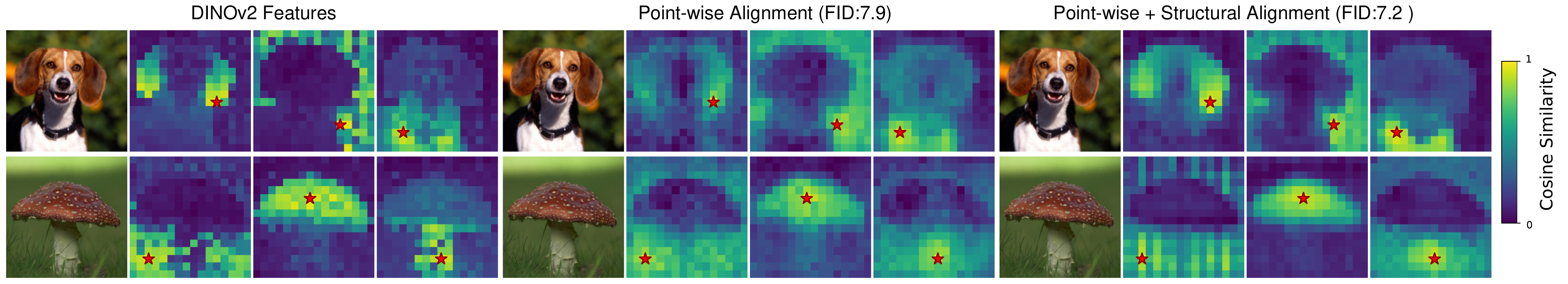} 
    \caption{\textbf{Effectiveness of Spatial Structural Supervision.} Comparison of similarity maps on the DINOv2 features and diffusion features of models trained with point-wise alignment and further integrated structural alignment. Point-wise only alignment is insufficient to mimic the token relationship of teacher features. Adding explicit structural supervision results in more concentrated token similarity and better image quality, confirming the effectiveness of structural supervision.}
    \label{fig:motivation_analysis}
\end{figure*}

\subsection{Motivation}
\label{subsec:motivation}
\paragraph{From Global Semantics to Spatial Structure.}
Building upon the flow matching framework, a series of works have achieved state-of-the-art generation performance. Nevertheless, training these models from scratch remains computationally intensive. To mitigate this inefficiency, the pioneer work REPA \citep{repa} aligns the noisy hidden projections of the denoising network with clean representations from pre-trained vision encoders and achieve performance boost. Despite empirical success to some extent, a fundamental research question~(RQ) raises:

\textit{RQ-1: What intrinsically drives the effectiveness of representation alignment—global semantics or spatial structure?}

Subsequently, iREPA \citep{irepa} provides a comprehensive analysis across 27 different vision encoders and uncovers an inspiring phenomenon: \textbf{spatial structure rather than global semantics, drives the generation performance intrinsically.} To learn the structural features from vision encoders, they introduce two straightforward architecture-level modifications: (1) replacing the commonly-used MLP in projection with a spatial convolution and (2) introducing spatial normalization to amplify the contrast between individual spatial tokens. These architectural tweaks help sharpen the spatial signal within the representations, achieving improved generation quality. This finding motivates us to dive deeper into the fundamental working mechanism of alignment:

\textit{RQ-2: If spatial structure is indeed the key, are commonly-used point-wise alignment capable of fully capturing it?}

\paragraph{The Limitation of Point-wise Alignment.}
We delve into alignment mechanics to understand the current alignment limitations. Let $ f $ and $ x_0 $ denote the pre-trained visual encoder (teacher) and a clean image, respectively. Let $\mathbf{H}^T=f(x_0) \in \mathbb{R}^{N \times D_T}$ denote the target visual representations, where $N, D_T$ are the number of tokens and the embedding dimension, respectively. Let $\mathbf{H}^S \in \mathbb{R}^{N \times D_S}$ represent the hidden states of the student, where $D_S$ denotes the diffusion embedding dimension. To align the feature dimensions, a trainable projection head $g_\phi$ is employed to map the student features to the teacher's latent space.

Most existing methods encourage feature alignment by maximizing the patch-wise similarity, which is equivalent to minimizing the negative cosine similarity averaged over all patches:
\begin{equation}
    \mathcal{L}_{\text{point}} = \frac{1}{N} \sum_{i=1}^N \left( 1 - \mathcal{S}_{\cos}(g_\phi(\mathbf{h}_i^S), \mathbf{h}_i^T) \right),
    \label{eq:point_loss}
\end{equation}
where $i$ denotes the token index, $\mathbf{h}_i^S, \mathbf{h}_i^T$ are the feature vectors of the $i$-th token, and $\mathcal{S}_{\cos}(\cdot, \cdot)$ is the cosine similarity function.

According to Eq.~\eqref{eq:point_loss}, we observe that this loss treats the feature map as an unstructured \textit{bag of vectors} and operates independently on each spatial location, enforcing local feature consistency while remaining agnostic to the relational topology between tokens. Even if the teacher model possesses a rich understanding of global geometry, the student is strictly penalized only for local deviations, ignoring the rich relational information for coherent image generation. So the point-wise alignment objective essentially falls short of capturing the spatial structural information, which is essential for coherent image generation.

\paragraph{Explicit Spatial Structural Supervision.}
We hypothesize that the bottleneck of prior works lies not in the teacher's representational capability but in the alignment protocol. To validate this, we conduct an experiment by explicit integrate structural supervision into the standard REPA training framework. Specifically, we retain the original point-wise alignment and introduce an auxiliary objective to force the student's pairwise token relationships (Gram matrix) to regress those of the teacher. We use the SiT-XL/2 model trained for 400k to exhibit similarity visualization results.

As shown in Figure~\ref{fig:motivation_analysis}, we find that the feature maps derived from pure point-wise alignment exhibit low contrast and diffuse attention. In contrast, introducing structural supervision produces more \textit{concentrated and meaningful} similarity maps with better generation FID ($7.9 \to 7.2$). This evidence confirms that introducing explicit structural supervision is essential for learning structural representations. Motivated by these observations, we formally propose the structural representation alignment method and detail its formulation in the following section.

\subsection{Structural Representation Alignment}
\label{subsec:srepa}
Based on the above analysis, we propose \textbf{sREPA}, a simple yet effective method for improving diffusion transformers by incorporating explicit structural supervision into the representation alignment process. We first normalize the features from the teacher and mapped features from the student to obtain $\mathbf{Z}^T$ and $\mathbf{Z}^S$ with L2-normalization. Then, the neighboring similarity distribution is extracted in each token map through the pairwise cosine similarity matrix $\mathbf{S} \in \mathbb{R}^{N \times N}$: 
\begin{equation}
    S_{ij}^T = \mathbf{z}_i^T \cdot (\mathbf{z}_j^T)^\top, S_{ij}^S = \mathbf{z}_i^S \cdot (\mathbf{z}_j^S)^\top.
\end{equation}
After that, we can use the matrix for further structural representation alignment. To prevent trivial solutions where the model minimizes loss by simply predicting itself, we explicitly mask the diagonal elements, considering only the off-diagonal relationships where $i \neq j$. Based on the pairwise cosine similarity matrix, we propose the following two variants of sREPA to align the structure distributions between the student and the teacher.

\paragraph{Variant 1: sREPA-MSE.}
Our primary strategy is to directly regress the student's relational structure to imitate the teacher. We minimize the MSE loss between the masked similarity matrices of the student and those of the teacher:
\begin{equation}
    \mathcal{L}_{\text{struc}}^{\text{MSE}} = \frac{1}{N(N-1)} \sum_{i=1}^N \sum_{j \neq i} \| S_{ij}^T - S_{ij}^S \|^2.
\end{equation}
This formulation is \textit{parameter-free} and straightforward. By enforcing the student's Gram matrix to regress to the teacher's, sREPA-MSE ensures that the relative geometry of the latent space is well preserved.

\paragraph{Variant 2: sREPA-KL.}
Alternatively, structural alignment can be formulated as a distribution matching problem. Instead of regressing raw similarity values, we view the row $\mathbf{S}_{i\cdot}$ as a probability distribution over the sequence.
We apply the Softmax function to the off-diagonal similarities to obtain the attention distribution for each token $i$ and $j$:
\begin{equation}
    P_{ij}^T = \frac{\exp(S_{ij}^T / \tau_t)}{\sum_{k \neq i} \exp(S_{ik}^T / \tau_t)}, \quad 
    P_{ij}^S = \frac{\exp(S_{ij}^S / \tau_s)}{\sum_{k \neq i} \exp(S_{ik}^S / \tau_s)},
    \label{eq:temperature}
\end{equation}
where $j \neq i$, $\tau_t$ and $\tau_s$ are temperature parameters controlling the sharpness of the distributions. 
The alignment is then enforced via the Kullback-Leibler (KL) divergence averaged over all tokens:
\begin{equation}
    \mathcal{L}_{\text{struc}}^{\text{KL}} = \frac{1}{N} \sum_{i=1}^N \sum_{j \neq i} P_{ij}^T \log \left( \frac{P_{ij}^T}{P_{ij}^S} \right).
\end{equation}
This variant emphasizes matching the relative importance of neighbors rather than absolute values. Temperatures $\tau_s $ and $\tau_t$ are employed to modulate the entropy of the relational distribution, which we will analyze in Section~\ref{sec:exp}.

\subsection{Training Objective}
\label{subsec:total_loss}
The training objective in most of our experiments is formulated as a weighted sum of the standard flow matching loss, the point-wise projection loss, and our proposed structural alignment loss:
\begin{equation}
    \mathcal{L}_{\text{total}} = \mathcal{L}_{\text{flow}} + \lambda_{proj} \mathcal{L}_{\text{proj}} + \lambda_{struc} \mathcal{L}_{\text{struc}},
\end{equation}
where $\mathcal{L}_{\text{struc}}$ represents either the $\mathcal{L}_{\text{struc}}^{\text{MSE}}$ or $\mathcal{L}_{\text{struc}}^{\text{KL}}$ variant.
The scalar $\lambda_{proj}$ and $\lambda_{struc}$ govern the relative importance of the two alignment losses.
We provide comprehensive ablations on the choice of structural loss variants and the sensitivity of these balancing weights in Section~\ref{subsec:ablation}. And in Section~\ref{subsec:Compatibility}, we further verify the compatibility of our sREPA with exisiting works.
Overall, the alignment procedure is summarized in Algorithm~\ref{alg:srepa_alignment} in the Appendix.

\section{Experiments}
\label{sec:exp}
In this section, we conduct comprehensive experiments to validate the effectiveness of our sREPA. In particular, we mainly investigate the following research questions:
\begin{itemize}
    \item \textbf{Effectiveness:} Can adding explicit structural guidance improve generation quality and convergence speed of diffusion transformers compared to standard point-wise alignment methods? (Sec.~\ref{subsec:main_results})
    \item \textbf{Robustness:} Does sREPA consistently maintain its performance advantage across varying model scales and diverse teacher representations? (Sec.~\ref{subsec:ablation})
    \item \textbf{Compatibility:} Can structure loss serve as a plug-and-play module that provides consistent improvements with other training pipelines? (Sec.~\ref{subsec:Compatibility})
\end{itemize}

\subsection{Setup}
\label{subsec:setup}
\textbf{Implementation Details.}
We follow the training protocols of REPA~\citep{repa} throughout all experiments. Specifically, we use ImageNet~\citep{imagenet}, where each image is preprocessed to the resolution of $256 \times 256$ and follows ADM~\citep{adm} for other data preprocessing protocols. Each image is then encoded into latent representations $\mathbf{z} \in \mathbb{R}^{32 \times 32 \times 4}$ using the pre-trained Stable Diffusion VAE~\citep{ldm}.
The training batch size is 256 with Exponential Moving Average (EMA) following REPA~\citep{repa}. 

\textbf{Evaluation Protocol.}
To systematically evaluate generation quality, we compute a suite of standard quantitative metrics on 50{,}000 generated samples. We report Fréchet Inception Distance (FID)~\citep{fid}, spatial FID (sFID)~\citep{sfid}, Inception Score (IS)~\citep{IS}, Precision and Recall ~\citep{recall}. All metrics are computed using the ADM evaluation protocol, ensuring comprehensive comparison with existing works. Unless otherwise specified, all samples are generated using Euler-Maruyama sampler with 250 steps and without classifier-free guidance~\citep{ho2022classifier}. More setup details about training and evaluation are provided in the Appendix~\ref{appendix-implementation}.

\subsection{Main Results}
\label{subsec:main_results}
\textbf{Accelerating Training Convergence with Improved Performance.}
As shown in Table~\ref{tab:wo_cfg}, sREPA consistently improves generation quality across all model scales~(B, L, XL). On the smaller SiT-B/2 backbone, sREPA significantly outperforms the REPA baseline, achieving an FID of 20.5 at 400K iterations. On the SiT-XL/2 scale, sREPA reaches an FID of \textbf{7.17} at 400K steps, outperforms REPA (7.9) and iREPA (7.52) counterparts, even surpassing the vanilla SiT (8.3) trained for 7M steps. Furthermore, sREPA outperforms REPA up to \textbf{1M} iterations, which means our method is not only an early-stage accelerator but consistently remains beneficial in longer training iterations.

\begin{table*}[t]
\centering

\begin{minipage}{0.46\linewidth}
\centering
\captionof{table}{\textbf{FID comparison on ImageNet 256×256.} All results are reported without classifier-free guidance (CFG).}
\label{tab:wo_cfg}
\resizebox{\linewidth}{!}{
\setlength{\tabcolsep}{12pt}
\begin{tabular}{lccc}
\toprule
Model & \#Params & Iter. & FID$\downarrow$  \\
\midrule
SiT-B/2 & 130M & 400K & 33.0 \\
+ REPA & 130M & 400K & 24.4 \\
{\textbf{+ sREPA(Ours)}} & 130M & \textbf{400K}  & \textbf{20.5}\\
\midrule
SiT-L/2 & 458M & 400K & 18.8 \\
+ REPA & 458M & 400K & 9.7 \\
{\textbf{+ sREPA(Ours)}} & 458M & \textbf{400K}  & \textbf{8.73}\\
\midrule
SiT-XL/2 & 675M & 400K & 17.2\\
SiT-XL/2 & 675M & 7M & 8.3\\
+ REPA & 675M & 200K   & 11.1 \\
+ REPA & 675M & 400K   & 7.9 \\
+ REPA & 675M & 1M   & 6.4 \\
+ iREPA & 675M & 400K  & 7.52 \\
+ iREPA & 675M & 1M  & 6.67 \\
{\textbf{+ sREPA(Ours)}}  & 675M & \textbf{200K}  & \textbf{9.36}\\
{\textbf{+ sREPA(Ours)}}  & 675M & \textbf{400K}  & \textbf{7.17}\\
{\textbf{+ sREPA(Ours)}}  & 675M & \textbf{1M}  & \textbf{6.15}\\
\bottomrule
\end{tabular}}
\end{minipage}
\hfill
\begin{minipage}{0.50\linewidth}
\centering
\captionof{table}{\textbf{Comprehensive comparison of different methods on ImageNet 256×256 with classifier-free guidance (CFG).}}
\label{tab:cfg}
\resizebox{\linewidth}{!}{
\setlength{\tabcolsep}{4pt}
\begin{tabular}{l c c c c c c}
\toprule
Model & Epochs  & FID$\downarrow$ & sFID$\downarrow$ & IS$\uparrow$ & Pre.$\uparrow$ & Rec.$\uparrow$ \\
\midrule
\multicolumn{7}{l}{\emph{Pixel diffusion}} \\
ADM-U  &400 & 3.94 & 6.14 & 186.7 & 0.82 & 0.52 \\
VDM$++$ & 560 & 2.40 & - & 225.3 & - & - \\
Simple diffusion & 800 & 2.77 & - & 211.8 & - & - \\
CDM & 2160 & 4.88 & - & 158.7 & - & - \\
\cmidrule(lr){1-7}
\multicolumn{7}{l}{\emph{Latent diffusion}} \\
LDM-4 & 200  & 3.60 & - & 247.7 & 0.87 & 0.48 \\
U-ViT-H/2 & 240 & 2.29 & 5.68  & 263.9 & 0.82 & 0.57 \\
MaskDiT & 1600 & 2.28 & 5.67 & 276.6 & 0.80 & 0.61 \\ 
SD-DiT & 480 & 3.23 & - & - & - & - \\
\cmidrule(lr){1-7}
DiT-XL/2 & 1400 & 2.27 & 4.60 & \textbf{278.2} & 0.83 & 0.57  \\
\cmidrule(lr){1-7}
SiT-XL/2 & 1400 & 2.06 & 4.50 & 270.3 & 0.82 & 0.59 \\
+ REPA & 80 & 2.39 & 4.64 & 246.8 & 0.82 & 0.57 \\
+ REPA & 200 & 1.96 & \textbf{4.49} & 264.0 & 0.82 & 0.60 \\
\textbf{+ sREPA(Ours)} & 80 & 2.25 & 4.65 & 257.5 & 0.83 & 0.58 \\
\textbf{+ sREPA(Ours)} & 200 & \textbf{1.91} & 4.50 & 271.8 & \textbf{0.83} & \textbf{0.60} \\
\bottomrule
\end{tabular}}
\end{minipage}
\end{table*}

\begin{figure*}[t]
    \centering
    \includegraphics[width=\linewidth]{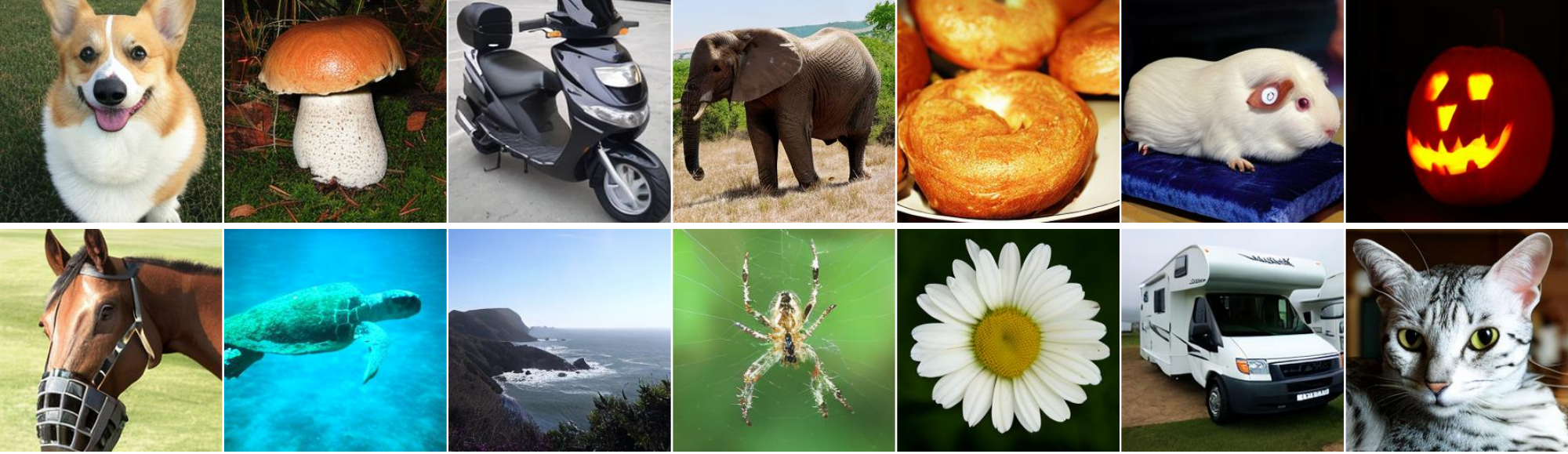} 
    \caption{Samples are generated on ImageNet $256\times256$ with the sREPA~(SiT-XL/2 model). Classifier-free guidance is applied with scale $w = 4.0$.}
    \label{fig:cfg_visualization}
\end{figure*}

\textbf{Comparison with State-of-the-art Methods.}
Table~\ref {tab:cfg} presents a comprehensive comparison against recent state-of-the-art methods with applying classifier-free guidance. Our framework achieves better performance than other methods, highlighting the superior convergence property. We also provide qualitative results of SiT-XL/2 with sREPA in Figure~\ref{fig:cfg_visualization}, which demonstrate the high-quality generation by applying the structural supervision. We provide more examples to illustrate the superior generation capability of our sREPA in Appendix~\ref{appendix-samples}.

\subsection{Ablation Study}
\label{subsec:ablation}
In this section, we conduct a systematic analysis to evaluate each component in sREPA. Unless otherwise specified, experiments are performed on SiT-B/2 model with DINOv2-B as target representation.

\begin{table*}[t]
\centering
\begin{minipage}{0.48\linewidth}
\centering
\caption{\textbf{Effectiveness of Structural Supervision.} Injecting either structural supervision objective outperforms the point-wise baseline.}
\label{tab:loss-variants}
\resizebox{\linewidth}{!}{
\setlength{\tabcolsep}{8pt}
\begin{tabular}{lcccccc}
\toprule
Method & $\mathcal{L}_{proj}$ & $\mathcal{L}_{struc}$ & Iter. & FID$\downarrow$ & sFID$\downarrow$ & IS$\uparrow$ \\
\midrule
\multicolumn{7}{l}{\textit{w/o Point-wise Alignment}} \\
sREPA & -- & MSE & 100K & 51.4 & 6.82 & 26.6\\
sREPA & -- & KL  & 100K & 48.2 & 6.65 & 28.4 \\
\cmidrule(lr){1-7}
sREPA & -- & MSE & 400K & 25.6 & 6.30 & 58.1 \\
sREPA & -- & KL  & 400K & 24.3 & 6.35 & 60.5 \\
\midrule
\multicolumn{7}{l}{\textit{w/ Point-wise Alignment}} \\
REPA & \checkmark & -- & 100K & 49.5 & 7.00 & 27.5 \\
sREPA & \checkmark & MSE & 100K & \textbf{41.3} & \textbf{6.71} & \textbf{35.1} \\
sREPA & \checkmark & KL & 100K & 43.4 & 6.94 & 33.1 \\
\cmidrule(lr){1-7}
REPA & \checkmark & -- & 400K & 24.4 & 6.40 & 59.9 \\
sREPA & \checkmark & MSE & 400K & \textbf{21.4} & \textbf{6.40} & 67.8 \\
sREPA & \checkmark & KL & 400K & 21.8 & 6.48 & \textbf{68.2} \\
\bottomrule
\end{tabular}
}
\end{minipage}
\hfill
\begin{minipage}{0.48\linewidth}
\centering
\caption{\textbf{Temperature Analysis.} We train all the SiT-B/2 models for 100k iterations here.}
\label{tab:ablation-temperature}
\resizebox{\linewidth}{!}{
\begin{small}
\setlength{\tabcolsep}{6pt}
\begin{tabular}{l c c c c c}
\toprule
Experiment & $\tau_t$ & $\tau_s$ & FID$\downarrow$ & sFID$\downarrow$ & IS$\uparrow$ \\
\midrule
REPA & - & - & 49.5 & 7.00 & 27.5 \\
\midrule
\multicolumn{6}{l}{\textit{I. Temperature Match (Fix $\tau_t=0.2$)}} \\
\multirow{4}{*}{Asymmetric}
  & 0.20 & 0.15 & 42.5 & 6.71 & 33.1 \\
  & 0.20 & 0.20 & \textbf{42.1} & 6.60 & \textbf{33.3} \\
  & 0.20 & 0.40 & 42.5 & \textbf{6.57} & 33.2 \\
  & 0.20 & 0.60 & 43.7 & 6.92 & 32.4 \\
\midrule
\multicolumn{6}{l}{\textit{II. Global Magnitude Search ($\tau_t = \tau_s$)}} \\
\multirow{4}{*}{Symmetric} 
  & 0.10 & 0.10 & 43.5 & 6.61 & 31.7 \\
  & 0.20 & 0.20 & \textbf{42.1} & \textbf{6.60} & \textbf{33.3} \\
  & 0.30 & 0.30 & 42.9 & 6.88 & 33.0 \\
  & 0.50 & 0.50 & 44.2 & 7.04 & 32.0 \\
\bottomrule
\end{tabular}
\end{small}
}
\end{minipage}
\vskip -0.1in
\end{table*}

\textbf{Structural Loss Variants.} 
To verify the necessity of explicit structural supervision, we evaluate the two $\mathcal{L}_{struc}$ variants: MSE-based and KL-based objectives. For a controlled comparison, we maintain the weight of point-wise alignment loss identical to REPA baseline, isolating the impact of the added structural constraints. As shown in Table~\ref{tab:loss-variants}, incorporating structural supervision consistently outperforms the baseline regardless of the specific loss formulation. Comparing the two variants, we observe that MSE and KL yield comparable performance gains, with MSE variant exhibiting a marginal advantage in final generation quality. We guess that MSE directly enforces alignment on raw similarity magnitudes, avoiding the potential signal dilution caused by the softmax normalization in the KL objective. Consequently, we adopt MSE variant as the default setting for sREPA.

Notably, relying solely on structural supervision without point-wise alignment fails to improve the baseline (Table~\ref{tab:loss-variants} top rows). We claim that \textit{point-wise alignment serves as a necessary semantic anchor that grounds features in the absolute latent space; without it, the structural constraints become underdetermined and ambiguous}. Our structural supervision is a complement to the previous point-wise alignment. 

\textbf{Temperature Analysis.}
The KL-based variant utilizes temperature ($\tau_s, \tau_t$) to modulate the entropy of the relational distribution (Eq.~\ref{eq:temperature}). We investigate the optimal configuration through a two-stage empirical strategy: (I) verifying the symmetric temperatures, and (II) searching for the optimal global temperature. The results are reported in Table~\ref{tab:ablation-temperature}. 

\textit{(1) Temperature Matching.}
Prior literature~\citep {ISD,CMD} employs an asymmetric temperature strategy ($\tau_t < \tau_s$) to obtain high-confidence predictions. However, in Table~\ref{tab:ablation-temperature} (Part I), we observe that this asymmetry is counter-productive for generation, indicating a fundamental difference from previous discriminative tasks: structural alignment for generation requires the student to faithfully replicate the teacher's relational topology. Symmetric temperatures ($\tau_s = \tau_t$) ensure that the relative distance ratios between tokens are preserved without distortion.

\textit{(2) Global Temperature Search.}
Fixing the symmetric constraint, we search for the optimal global temperature. As shown in Table~\ref{tab:ablation-temperature} (Part II), $\tau_s = \tau_t = 0.2$ achieves the best overall performance. Increasing the temperature further leads to feature distribution over-smoothing, diluting the structural signal. Conversely, an excessively low temperature drives the distribution towards a nearly Dirac delta distribution, weakening the ability to perceive structural relationships. Consequently, we adopt $\tau_s = \tau_t = 0.2$ as the default setting for the KL variant.

\subsection{Generalization Capability Evaluation}
We further conduct comprehensive experiments to explore the generalization capability of our sREPA in terms of different alignment depths, diverse vision encoders, and compatibility with other methods.

\begin{wraptable}{r}{0.46\linewidth}
\vspace{-15pt}
\centering
\caption{\textbf{Ablation study for Alignment Depth.} All experiments are conducted on SiT-B/2, trained for 100K iterations.}
\vspace{-5pt}
\label{tab:ablation-depth}
\resizebox{\linewidth}{!}{
\begin{small}
\setlength{\tabcolsep}{6pt}
\begin{tabular}{l c c c c c}
\toprule
Aln. Depth & FID $\downarrow$ & sFID $\downarrow$ & IS $\uparrow$ & Pre.$\uparrow$ & Rec.$\uparrow$ \\
\midrule
Layer 4 & 46.2 & 7.00 & 30.4 & 0.47 & 0.61 \\
\textbf{sREPA} & \textbf{39.3} & \textbf{6.74} & \textbf{37.8} & \textbf{0.51} & \textbf{0.61} \\
\midrule
Layer 6 & 47.1 & 7.25 & 30.1 & 0.48 & 0.60 \\
\textbf{sREPA} & \textbf{41.2} & \textbf{7.11} & \textbf{36.3} & \textbf{0.50} & \textbf{0.61} \\
\midrule
Layer 8 & 51.6 & 8.07 & 27.8 & 0.45 & 0.59 \\
\textbf{sREPA} & \textbf{45.5} & \textbf{7.58} & \textbf{32.9} & \textbf{0.48} & \textbf{0.60} \\
\bottomrule
\end{tabular}
\end{small}}
\vspace{-20pt}
\end{wraptable}
\textbf{Alignment Depth.}
We first investigate the generalization capability of our approach across different alignment depths~(Layer 4, 6, 8), compared with REPA. As shown in Table~\ref{tab:ablation-depth}, our method demonstrates consistent improvements over the REPA baseline by substantial gains across all depths. We attribute these gains to the explicit spatial structure alignment, which effectively enforces geometric consistency between the student and teacher representations and eventually improves generation quality.

\textbf{Diverse Vision Encoders.}
To verify that sREPA serves as a robust, encoder-agnostic enhancer for diffusion transformers, we evaluate the performance across a wide range of teacher representations.
Specifically, we employ discriminative models DINOv2~\citep{dinov2}, DINOv3~\citep{dinov3}, MoCoV3~\citep{chen2021empirical}, multi-modal model CLIP~\citep{clip}, reconstructive model MAE~\citep{mae}, and joint-embedding predictive model I-JEPA~\citep{assran2023self} as teacher encoders. As shown in Table~\ref{tab:generalization-encoder}, sREPA yields consistent and significant performance gains over the vanilla SiT baseline across all tested encoders. By synergizing point-wise alignment with distribution-wise coherence, sREPA fully leverages the diverse semantic and structural information embedded within these encoders. This mechanism ensures the effective transfer of rich representations to the diffusion backbone, significantly enhancing generation quality and proving its effectiveness in standard training scenarios.

\begin{table*}[t]
\centering
\begin{minipage}{0.46\linewidth}
\centering
\caption{\textbf{Generalization across Diverse Vision Encoders.} We employ diverse teacher encoders to demonstrate that sREPA consistently enhances generation fidelity regardless of the teacher's representation variants. All experiments are conducted using SiT-B/2 trained for 400k iterations.}
\label{tab:generalization-encoder}
\resizebox{\linewidth}{!}{
\setlength{\tabcolsep}{4pt}
\begin{tabular}{l c c c c c c}
\toprule
Method & Encoder & FID$\downarrow$ & sFID$\downarrow$ & IS$\uparrow$ & Pre.$\uparrow$ & Rec.$\uparrow$ \\
\midrule
SiT-B/2 & -- & 33.0 & 6.46 & 43.7 & 0.53 & 0.63 \\
\midrule
\multirow{6}{*}{sREPA}
  & MAE-L    & 28.3 & \textbf{6.12} & 51.8 & 0.57 & 0.64 \\
  & MoCoV3-B & 22.9 & 6.41 & 62.5 & 0.60 & 0.64 \\
  & I-JEPA-H & 22.2 & 6.65 & 67.8 & 0.60 & 0.65 \\
  & CLIP-L   & 21.0 & 6.41 & 69.1 & 0.61 & 0.64 \\
  & DINOv3-B & 22.1 & 6.54 & 67.6 & 0.59 & 0.65 \\
  & DINOv2-B & \textbf{20.5} & 6.53 & \textbf{71.2} & \textbf{0.61} & \textbf{0.65} \\
\bottomrule
\end{tabular}}
\end{minipage}
\hfill
\begin{minipage}{0.50\linewidth}
\centering
\caption{\textbf{Compatibility with existing frameworks.} We integrate sREPA into existing framework with training from scratch. All experiments are conducted on SiT-B/2 with DINOv2-B as the target representation. Results demonstrate consistent improvements across all metrics, verifying the compatibility of sREPA.}
\label{tab:compatibility}
\resizebox{\linewidth}{!}{
\begin{small}
\setlength{\tabcolsep}{8pt}
\begin{tabular}{l c c c c c c}
\toprule
Method & Iter & FID $\downarrow$ & sFID $\downarrow$ & IS $\uparrow$ & Pre. $\uparrow$ & Rec. $\uparrow$ \\
\midrule
iREPA & 400K & 21.4 & 6.77 & 70.9 & 0.60 & \textbf{0.65} \\
\hspace{0.5em} \textbf{+ Ours} & 400K & \textbf{20.3} & \textbf{6.70} & \textbf{73.4} & \textbf{0.61} & 0.64 \\
\midrule
REG & 100K & 36.1 & 7.74 & 45.5 & 0.53 & 0.61 \\
\hspace{0.5em} \textbf{+ Ours} & 100K & \textbf{30.9} & \textbf{7.24} & \textbf{53.4} & \textbf{0.57} & \textbf{0.61} \\
\cmidrule(lr){1-7}
REG & 200K & 22.1 & 7.19 & 72.2 & 0.60 & \textbf{0.63} \\
\hspace{0.5em} \textbf{+ Ours} & 200K & \textbf{19.8} & \textbf{6.96} & \textbf{79.0} & \textbf{0.62} & 0.62 \\
\cmidrule(lr){1-7}
REG & 400K & 15.2 & \textbf{6.63} & 94.6 & 0.64 & 0.63 \\
\hspace{0.5em} \textbf{+ Ours} & 400K & \textbf{14.4} & 6.94 & \textbf{98.0} & \textbf{0.65} & \textbf{0.63} \\
\bottomrule
\end{tabular}
\end{small}
}
\end{minipage}
\end{table*}

\textbf{Compatibility Study.}
\label{subsec:Compatibility}
A key strength of our structural representation alignment is the orthogonality to the standard point-wise alignment objective. Existing frameworks primarily focus on refining point-to-point feature regression and overlook geometric relationships within the feature manifold. Consequently, sREPA can function as a \textbf{complementary plug-and-play module}, injecting explicit structural guidance into existing pipelines without conflicting with their original objectives.

To demonstrate this versatility, we integrate sREPA into another two distinct representation alignment frameworks that rely exclusively on point-wise alignment:
\begin{itemize}
    \item \textbf{iREPA}~\citep{irepa}: An improved version of REPA that replaces the MLP projector with a convolutional layer and incorporates spatial normalization for target patch tokens.
    \item \textbf{REG}~\citep{reg}: A framework that provides discriminative guidance to the model by entangling image latents with the foundation model's class token.
\end{itemize}
We evaluate compatibility by injecting our structural alignment loss (using the MSE loss variant) into their original objectives and training from scratch. As detailed in Table~\ref{tab:compatibility}, integrating sREPA consistently boosts the performance of both baselines. Specifically, for iREPA, the integration of structural supervision reduces FID from 21.4 to 20.3. 
For the stronger baseline REG, which achieves a remarkable performance (FID: 15.2, IS: 94.6) at 400k training iterations, injecting our method yields further gains (FID: 14.4, IS: 98.0). Consistent gains confirm that sREPA captures meaningful geometric dependencies neglected by point-wise constraints inherently. By seamlessly injecting structural supervision, sREPA effectively refines the student's representational manifold, achieving superior generative fidelity without any architectural modifications.

\section{Conclusion}
In this paper, we have presented sREPA, an efficient and compatible structure alignment framework for improving diffusion training. Particularly, we validate that the commonly used point-wise alignment is insufficient for explicitly capturing the spatial structure in vision foundation models. We have shown incoporating explicit structure-aware supervision into different methods can consistently improve the generation performance over the original point-wise alignment method, demonstrating that explicit structural alignment substantially improves convergence speed and generation quality.

{
\small
\bibliographystyle{plainnat} \bibliography{neurips_2026}
}

\clearpage
\newpage

\appendix

\section{Implementation Details}
\label{appendix-implementation}
\subsection{Training Details}
\label{appendix-train}
We follow the same experimental setup as in REPA~\citep{repa}. All training experiments are conducted on the ImageNet~\citep{imagenet} training split. For preprocessing, we adopt the protocol from the ADM framework in which images are center-cropped and resized to 256 × 256 resolution. For optimization, we use AdamW~\citep{kingma2014adam,loshchilov2017decoupled} with a constant learning rate of $1 \times 10^{-4}$, $(\beta_1, \beta_2) = (0.9, 0.999)$, and no weight decay. The global batch size is set to 256. During training, we use bfloat16 mixed precision and exponential moving average (EMA) for stable optimization. Following REPA~\citep{repa}, we use MLP with SiLU activations as the projection.
For REG~\citep{reg} and iREPA~\citep{irepa} experiments, we use the official open-source implementation.

\subsection{Evaluation Details}
\label{appendix-eval}
For image generation evaluation, we strictly follow the ADM setup~\citep{adm}. We report generation quality using Fr\'echet Inception Distance (gFID)~\citep{fid}, structural FID (sFID)~\citep{sfid}, inception score (IS)~\citep{IS}, precision (Pre.) and recall (Rec.) measured on 50K generated images. For sampling, we use the same SDE Euler-Maruyama sampler in REPA and sample for 250 steps. For image decoding, we use the \texttt{stabilityai/sd-vae-ft-ema} decoder to convert latent representations into images.

\subsection{Hyperparameters and Computation Resources}
\label{appendix-hyperparameters}
We report the hyperparameters for architecture, interpolants, optimization, and our sREPA experiment settings in Table ~\ref{tab:hyper}. ``EM'' denotes the Euler--Maruyama solver. All training and evaluation experiments are conducted on a single machine with 8 NVIDIA H200 GPUs.

\section{Algorithm of sREPA}
\label{appendix-algorithm}
In this section, we provide the alignment procedure of sREPA in Algorithm~\ref{alg:srepa_alignment}. The algorithm summarizes how the flow matching loss, projection loss, and structural alignment loss are computed during training.

\begin{algorithm}[b]
\caption{PyTorch-style Pseudocode of sREPA.}
\label{alg:srepa_alignment}
\definecolor{codeblue}{rgb}{0.25,0.5,0.5}
\definecolor{codekw}{rgb}{0.85, 0.18, 0.50}
\lstset{
  backgroundcolor=\color{white},
  basicstyle=\fontsize{7.2pt}{7.2pt}\ttfamily\selectfont,
  columns=fullflexible,
  breaklines=true,
  captionpos=b,
  commentstyle=\fontsize{7.2pt}{7.2pt}\color{codeblue},
  keywordstyle=\fontsize{7.2pt}{7.2pt},
  escapechar={|},
}
\begin{lstlisting}[language=python]
# z, z_til: teacher, student features
# lam_proj, lam_struc: loss weights
# tau_s, tau_t: temperatures for student/teacher

def sREPA_loss(z, z_til, mode='MSE'):
    # 1. Point-wise Alignment
    z = F.normalize(z, dim=-1)
    z_til = F.normalize(z_til, dim=-1)
    loss_proj = -(z * z_til).sum(dim=-1).mean()

    # 2. Structural Alignment
    A = torch.bmm(z, z.transpose(1, 2))
    A_til = torch.bmm(z_til, z_til.transpose(1, 2))
    mask = torch.eye(A.shape[1]).bool()
    A = A[:, ~mask].reshape(A.shape[0], A.shape[1], A.shape[1] - 1)
    A_til = A_til[:, ~mask].reshape(A_til.shape[0], A_til.shape[1], A_til.shape[1] - 1)
    
    # 3. Loss Calculation
    if mode == 'MSE':
        loss_struc = F.mse_loss(A, A_til)
    elif mode == 'KL':
        P_t = F.softmax(A.detach() / tau_t, dim=-1)
        P_s = F.log_softmax(A_til / tau_s, dim=-1)
        loss_struc = F.kl_div(P_s, P_t, reduction='batchmean')
    return lam_proj * loss_proj + lam_struc * loss_struc
\end{lstlisting}
\end{algorithm}

\begin{table*}[t]
    \centering
    \caption{Hyperparameter settings across different model scales.}
    \resizebox{\linewidth}{!}{
    \setlength{\tabcolsep}{18.0pt}
    \begin{tabular}{lccccc}
        \toprule
        \textbf{Backbone} & SiT-B & SiT-L & SiT-XL &  \\
        \midrule
        \textbf{Architecture} \\
        \#Params  &130M    &458M    &675M \\
        Input & 32$\times$32$\times$4 & 32$\times$32$\times$4 & 32$\times$32$\times$4  \\
        Layers  & 12 & 24 & 28  \\
        Hidden dim.  & 768 & 1,024 & 1,152  \\
        Num. heads  & 12 & 16 & 16  \\ 
        \midrule
        \textbf{Interpolants} \\
        $\alpha_t$  & $1-t$ & $1-t$ & $1-t$ \\
        $\sigma_t$  & $t$ & $t$ & $t$  \\
        $w_t$  & $\sigma_t$ & $\sigma_t$ & $\sigma_t$  \\
        Training obj.  & v-prediction & v-prediction & v-prediction \\
        Sampler  & EM & EM & EM  \\
        Sampling steps  & 250 & 250 & 250 \\
        \midrule
        \textbf{Optimization} \\
        Batch size  & 256 & 256 & 256 \\ 
        Optimizer  & AdamW & AdamW & AdamW \\
        lr  & 0.0001 &  0.0001 & 0.0001 \\
        $(\beta_1, \beta_2)$  & (0.9, 0.999) & (0.9, 0.999) & (0.9, 0.999) \\
        \midrule
        \textbf{sREPA-MSE settings} \\
        $\lambda_{proj}$  & 1.0 & 1.0 & 1.0  \\
        $\lambda_{struc}$ & 2.0 & 2.0 & 2.0  \\
        Alignment depth  & 4 & 8 & 8  \\
        Encoder   & DINOv2-B & DINOv2-B & DINOv2-B  \\
        \midrule
        \textbf{sREPA-KL settings} \\
        $\lambda_{proj}$  & 1.0 & 1.0 & 1.0  \\
        $\lambda_{struc}$ & 0.5 & 0.5 & 0.5  \\
        Alignment depth  & 4 & 8 & 8  \\
        Encoder   & DINOv2-B & DINOv2-B & DINOv2-B  \\
        \bottomrule
    \end{tabular}
    \label{tab:hyper}}
\end{table*}

\section{More Experiments}
\subsection{Qualitative Evaluation}
\begin{figure*}[t]
    \centering
    \includegraphics[width=\linewidth]{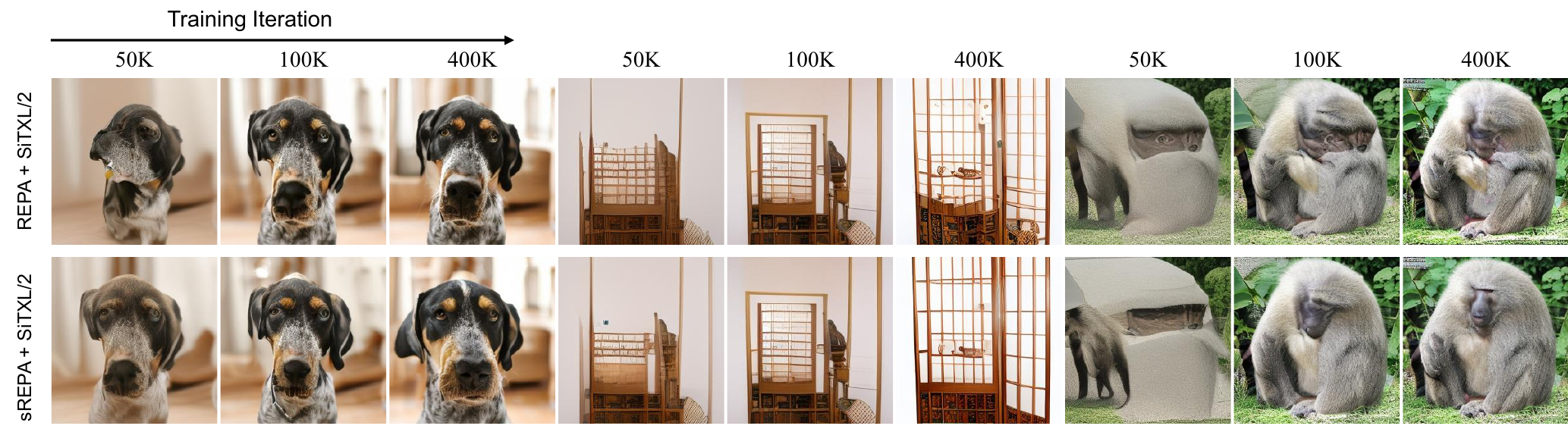} 
    \caption{sREPA improves visual scaling. We observe that sREPA produces higher-quality images at 400K steps compared with the vanilla-REPA and generates more structurally meaningful images even in the early stages of training. Results for both methods are sampled using the same seed, noises, and class labels.}
    \label{fig:compare}
\end{figure*}

We provide more qualitative comparisons between REPA and sREPA in Figure~\ref{fig:compare}. We generate images from the same noises and labels using checkpoints at 50K, 100K, and 400K training iterations, respectively. As shown in Figure~\ref{fig:compare}, sREPA generates more structurally meaningful images during early training stages, demonstrating superior image generation quality compared to the REPA baseline.

\subsection{Ablation Study on Loss Weights}
\label{appendix-lossweights}
We conduct more comprehensive ablations on the hyperparameters $\lambda_{proj}$ and $\lambda_{struc}$. Table~\ref{tab:appendix-lossweights.} details the optimization trajectory. All experiments are conducted on SiT-B/2 models trained at 400K iterations.

\textbf{MSE Variant.}
We first fix the $\lambda_{proj}=0.5$ and perform a parameter sweep over $\lambda_{struc}$ to assess the contribution of structural loss to the overall performance. As shown in Table~\ref{tab:appendix-lossweights.}, the integration of structural constraints leads to immediate performance gains, with the optimal results achieved at $\lambda_{struc}=2.0$ (FID: 20.9).
Subsequently, fixing $\lambda_{struc}=2.0$, we find that increasing $\lambda_{proj} = 1.0$ further improve generation quality (FID: 20.5). Further increasing the weight of the structural loss will not bring more generation performance. This indicates that the optimal configuration requires a strong structural penalty balanced by a moderate semantic anchor; neither component can be maximized in isolation without disrupting the equilibrium.

\textbf{KL Variant.}
Applying the same searching strategy to the KL formulation, we identify that the KL variant achieve best final performance when ($\lambda_{proj}=1.0, \lambda_{struc}=0.5$). 
Although the KL variant obtains a bit worse performance than the MSE variant, both variants outperform the base model SiT-XL and baseline model REPA. 

\begin{table*}[t]
    \centering
    \caption{\textbf{Ablation on Loss Weights.} To search for the best loss weights, we report results evaluated on SiT-B/2 trained for 400K iterations. \colorbox{gray!15}{Gray} highlights the optimum for each variant. }
    \resizebox{\linewidth}{!}{
    \label{tab:appendix-lossweights.}
        \setlength{\tabcolsep}{20pt}
        \begin{tabular}{l c c c c c}
            \toprule
            Method & $\lambda_{proj}$ & $\lambda_{struc}$ & FID $\downarrow$ & sFID $\downarrow$ & IS $\uparrow$ \\
            \midrule
            REPA & 0.5 & - & 24.4 & 6.40 & 59.9 \\
            \midrule
            \multirow{15}{*}{sREPA-MSE} 
              & 0.5 & 0.5 & 22.2 & 6.58 & 66.7\\
              & 0.5 & 1.0 & 21.7 & 6.62 & 68.1\\
              & 0.5 & 1.5 & 21.1 & 6.52 & 68.7\\
              & 0.5 & 2.0 & 20.9 & 6.51 & 69.7\\
              & 0.5 & 2.5 & 21.7 & 6.52 & 67.6\\
              & 0.5 & 3.0 & 21.2 & 6.52 & 69.0 \\
              & 0.5 & 3.5 & 21.5 & 6.53 & 68.4 \\
              & 0.5 & 4.0 & 21.4 & \textbf{6.41} & 67.8 \\
              \cmidrule{2-6}
              & 1.0 & 0.5 & 21.0 & 6.42 & 69.5 \\
              & 1.0 & 1.0 & 20.9 & 6.52 & 69.9 \\
              & 1.0 & 1.5 & 21.5 & 6.75 & 68.9 \\
              & \colorbox{gray!15}{1.0} & \colorbox{gray!15}{2.0} & \textbf{20.5} & 6.53 & \textbf{71.2} \\
              & 1.0 & 2.5 & 20.9 & 6.50 & 70.8 \\
              & 1.0 & 3.0 & 21.2 & 6.45 & 69.7 \\
              & 1.0 & 4.0 & 21.2 & 6.51 & 69.7 \\
            \midrule
            \multirow{12}{*}{sREPA-KL} 
              & 0.5 & 0.3 & 21.9 & \textbf{6.37} & 67.0 \\
              & 0.5 & 0.5 & 21.8 & 6.48 & 68.2 \\ 
              & 0.5 & 1.0 & 22.0 & 6.54 & 67.1 \\
              & 0.5 & 2.0 & 22.3 & 6.61 & 66.8 \\
              & 0.5 & 4.0 & 22.7 & 6.60 & 65.3 \\
              \cmidrule{2-6}
              & \colorbox{gray!15}{1.0} & \colorbox{gray!15}{0.5} & \textbf{21.4} & 6.47 & \textbf{68.5} \\
              & 1.0 & 1.0 & 21.5 & 6.44 & 68.4 \\
              & 1.0 & 1.5 & 21.4 & 6.42 & 68.2 \\
              & 1.0 & 2.0 & 22.0 & 6.40 & 66.4 \\
              & 1.0 & 2.5 & 21.5 & 6.41 & 67.1 \\
              & 1.0 & 3.0 & 21.9 & 6.58 & 66.7 \\
              & 1.0 & 4.0 & 22.0 & 6.43 & 66.9 \\
            \bottomrule
        \end{tabular}}
    \vskip -0.1in
\end{table*}

\subsection{Effect of Point-Wise Alignment Strength}
\label{appendix-alignment-strength}
In this section, we further analyze the effect of point-wise alignment strength in the proposed sREPA framework. Specifically, we increase the weight of the point-wise alignment objective in both REPA and sREPA to study whether stronger point-wise supervision alone can further improve performance.

\begin{table*}[t]
\centering
\caption{\textbf{Effect of point-wise alignment strength.} We increase the point-wise alignment strength in REPA and sREPA to demonstrate that the performance gain mainly comes from the structural supervision rather than stronger point-wise alignment. All experiments are conducted using SiT-B/2 trained for 1M iterations.}
\resizebox{\linewidth}{!}{
\label{tab:alignment_strength}
\setlength{\tabcolsep}{15pt}
\begin{tabular}{lcc|cc|cc|cc}
\toprule
 &  &  & \multicolumn{2}{c}{100K} & \multicolumn{2}{c}{400K} & \multicolumn{2}{c}{1M} \\
\cmidrule(lr){4-5}\cmidrule(lr){6-7}\cmidrule(lr){8-9}
Method & $\lambda_{\text{proj}}$ & $\lambda_{\text{struc}}$ 
& FID$\downarrow$ & IS$\uparrow$ 
& FID$\downarrow$ & IS$\uparrow$ 
& FID$\downarrow$ & IS$\uparrow$ \\
\midrule
REPA  & 1.0 & -- & 43.3 & 32.9 & 22.3 & 66.8 & 17.4 & 82.1 \\
REPA  & 3.0 & -- & 41.4 & 35.3 & 21.6 & 69.1 & 17.2 & 84.2 \\
\midrule
sREPA & 1.0 & 2.0 & \textbf{39.9} & 36.6 & \textbf{20.5} & \textbf{71.2} & \textbf{16.5} & \textbf{85.0} \\
sREPA & 2.0 & 2.0 & 40.0 & 36.2 & 21.0 & 69.8 & 16.8 & 84.8 \\
sREPA & 4.0 & 2.0 & 40.2 & \textbf{37.5} & 21.4 & 70.3 & 16.8 & 84.8 \\
\bottomrule
\end{tabular}}
\end{table*}

As shown in Table~\ref{tab:alignment_strength}, while increasing $\lambda_{\text{proj}}$ alone yields only limited gains (i.e., FID 17.4 $\rightarrow$ 17.2 at 1M), sREPA with $\lambda_{\text{proj}}=1.0, 2.0, 4.0$ all perform better than the REPA baselines across all iterations. Besides, further increasing $\lambda_{\text{proj}}$ within sREPA does not improve performance, and in fact slightly degrades it. These results suggest that the improvement cannot be explained solely by stronger point-wise supervision and structural supervision consistently provides additional and useful training signals beyond point-wise alignment.

\begin{table*}[t]
    \centering
    \caption{Detailed evaluation results of SiT-XL+sREPA at 1M iterations with different classifier-free guidance scale $w$.}
    \label{tab:appendix-cfg}
    \resizebox{\linewidth}{!}{
    \setlength{\tabcolsep}{12.0pt}
    \begin{tabular}{l c c c c c c c c}
        \toprule
        Model & \#Params & Iter. & $w$ & FID$\downarrow$ & sFID$\downarrow$ & IS$\uparrow$ & Prec.$\uparrow$ & Rec.$\uparrow$ \\
        \midrule
        SiT-XL/2 & 675M & 7M & 1.500 & 2.06 & 4.50 & 270.3 & 0.82 & 0.59 \\
        \midrule
        + sREPA & 675M & 1M & 1.300 & 1.90 & 4.51 & 263.2 & 0.81 & 0.61 \\
        + sREPA & 675M & 1M & 1.325 & 1.91 & 4.50 & 271.8 & 0.82 & 0.60 \\
        + sREPA & 675M & 1M & 1.350 & 1.94 & 4.51 & 280.4 & 0.82 & 0.60 \\
        + sREPA & 675M & 1M & 1.375 & 2.01 & 4.53 & 287.6 & 0.83 & 0.60  \\
        + sREPA & 675M & 1M & 1.400 & 2.10 & 4.55 & 295.6 & 0.83 & 0.59 \\
        \bottomrule
    \end{tabular}}
\end{table*}

\subsection{Ablation Study on CFG Scale}
We further conduct an ablation study about different classifier-free guidance scales with SiT-XL/2 + sREPA at 400K iterations. The results are shown in Table~\ref{tab:appendix-cfg}. In our evaluation with CFG, we choose CFG scale $w=1.325$ as the default setting.

\subsection{Experiments on ImageNet $512 \times 512$}
To further validate the scalability of sREPA with respect to the input image resolution, we conduct additional experiments on ImageNet $512 \times 512$. We strictly follow the setup used in our ImageNet $256 \times 256$ experiment except an input dimension. We conduct this experiment on SiT/XL-2 model with DINOv2-B as target representation.

\begin{table*}[t]
    \centering
    \caption{\textbf{System-level comparison} on ImageNet $512 \times 512$. We use CFG with $w=1.35$.}
    \label{tab:exp512}
    \resizebox{\linewidth}{!}{
        \setlength{\tabcolsep}{14.0pt}
        \begin{tabular}{l c c c c c c}
        \toprule
        {Model} & Epochs  &  {FID$\downarrow$} & {sFID$\downarrow$} & {IS$\uparrow$} & {Pre.$\uparrow$} & Rec.$\uparrow$ \\
        \midrule
        \multicolumn{7}{l}{\emph{Pixel diffusion}\vspace{0.02in}} \\
        VDM$++$ & - & 2.65 & - & 278.1 & - & - \\
        ADM-G, ADM-U & 400 & 2.85 & 5.86 & 221.7 & 0.84 & 0.53 \\
        Simple diffusion (U-Net) & 800 & 4.28 & - & 171.0 & - & - \\
        Simple diffusion (U-ViT, L) & 800 & 4.53 & - & 205.3 & - & - \\
        \cmidrule(lr){1-7}
        \multicolumn{7}{l}{\emph{Latent diffusion}\vspace{0.02in}} \\
        MaskDiT & 800 &  2.50 & 5.10 & 256.3 & 0.83 & 0.56 \\ 
        \cmidrule(lr){1-7}
        {DiT-XL/2} & 600 & 3.04 & 5.02 & 240.8 & 0.84 & 0.54  \\
        \cmidrule(lr){1-7}
        {SiT-XL/2} & 600 & 2.62 & 4.18 & 252.2 & 0.84 & 0.57 \\
        {{+ REPA}} & 80  & 2.44 & 4.21 & 247.3 & 0.84 & 0.56 \\
        {{+ REPA}} & 100 & 2.32 & \textbf{4.16} & 255.7 & 0.84 & 0.56 \\
        \textbf{+ sREPA(Ours)}  & 80  & 2.40 & 4.24 & 254.3 & 0.84 & 0.54 \\
        \textbf{+ sREPA(Ours)}  & 100 & \textbf{2.29} & 4.18 & \textbf{260.3} & \textbf{0.84} & \textbf{0.57} \\
        \bottomrule
        \end{tabular}}
\end{table*}

As shown in Table~\ref{tab:exp512}, sREPA continues to provide consistent improvements over REPA under the same training budget. In particular, at 100 epochs, sREPA achieves a FID of 2.29 compared to 2.32 with REPA, while improving IS from 255.7 to 260.3. Similarly, at 80 epochs, sREPA achieves a slightly better FID (2.40 vs. 2.44) and higher IS (254.3 vs. 247.3). These results suggest that explicit structural supervision remains effective at higher resolutions.

\section{Overhead Analysis}
\label{appen:overhead}
The proposed sREPA incorporates pairwise token matching with quadratic complexity in the token count, therefore, to better support our training efficiency claim, we measure the training throughput and peak GPU memory usage of REPA, sREPA and iREPA. All experiments are conducted using SiT-XL/2 models with the same batchsize setting on 8 Nvidia H200 GPUs.

\begin{table*}[t]
    \centering
    \caption{\textbf{Training throughput and peak GPU memory usage} of different methods on ImageNet at different resolutions.}
    \label{tab:throughput-memory}
    \resizebox{\linewidth}{!}{
    \setlength{\tabcolsep}{18.0pt}
    \begin{tabular}{l c c c}
        \toprule
        Method & Dataset & Throughput (img/s)$\uparrow$ & Peak GPU Memory (GB)$\downarrow$ \\
        \midrule
        REPA  & ImageNet $256 \times 256$ & 2044 & 27.3 \\
        iREPA & ImageNet $256 \times 256$ & 2200 & 26.5 \\
        sREPA & ImageNet $256 \times 256$ & 2008 & 27.3 \\
        \midrule
        REPA  & ImageNet $512 \times 512$ & 585 & 65.5 \\
        iREPA & ImageNet $512 \times 512$ & 520 & 63.0 \\
        sREPA & ImageNet $512 \times 512$ & 573 & 67.0 \\
        \bottomrule
    \end{tabular}}
\end{table*}

As detailed in Table~\ref{tab:throughput-memory}, the results show that the structure alignment loss introduces modest throughput overhead and minimal additional memory cost. Compared with the REPA baseline, sREPA reduces training throughput by only 1.77\% on ImageNet $256 \times 256$ and 2.05\% on ImageNet $512 \times 512$. The extra memory cost from the Gram matrix is also small: 0.04 GB at 256 resolution and 1.50 GB at 512 resolution, corresponding to only 0.16\% and 2.30\% increases in peak GPU memory, respectively. These results indicate that the improvement in convergence speed and generation quality remains meaningful in practice, and the trade-off is worthful for diffusion transformer training.

\section{Limitations and Future Work}
\label{appen:limitation}
Despite the improvements achieved by our method, several limitations remain to be
addressed. First, although the overhead analysis in Table~\ref{tab:throughput-memory} shows that sREPA introduces only modest throughput and memory overhead on ImageNet $256\times256$ and $512\times512$, the structural alignment loss still relies on pairwise token matching with quadratic complexity in the number of tokens. Therefore, its computational cost may become more pronounced when applied to substantially larger token grids, higher-resolution generation, video generation, or other settings with long spatio-temporal sequences.
In such scenarios, the pairwise similarity matrix may become a non-negligible bottleneck in both computation and memory. Future work could investigate more scalable structural alignment strategies, such as local relation matching to reduce quadratic overhead while preserving the benefits of structural supervision.

Second, although we validate the effectiveness of sREPA on ImageNet $256\times256$ and $512\times512$, the current empirical scope remains relatively limited. Due to computational constraints, we do not extend our experiments to larger-scale text-to-image generation or long-video generation tasks. In such settings, accelerating the training of diffusion transformers remains an important and challenging problem. Integrating structural supervision to these more complex generative scenarios is an interesting direction for future work.


\section{Societal Impacts}
\label{appen:societal-impacts}
\subsection{Positive Impacts}
The problem of accelerating diffusion transformer training is practically important. sREPA can improve both convergence speed and final generation quality for the training of diffusion transformers by introducing explicit structural supervision into representation alignment. This may benefit a range of applications built upon visual generative models, such as content creation, data synthesis, and visual asset generation, where high-quality image generation is required. Furthermore, since sREPA is implemented as a loss-level modification and does not require architectural redesign, it can be integrated into existing representation alignment frameworks seamlessly. This could lower the barrier for training diffusion transformer, and may also reduce the energy consumption for training large-scale diffusion models.

\subsection{Negative Impacts}
Despite these benefits, sREPA may also introduce potential societal risks. By improving the quality and training efficiency of image generators, sREPA makes realistic synthetic images easier to produce. Such capability could be misused to create misleading visual content, fabricated media, or other forms of visual misinformation, raising privacy and ethical concerns.

\section{Declaration of LLM usage}
\label{appen:llm-usage}
In this work, LLMs were employed only for language-related tasks, including grammar correction, spelling checks, and word choice refinement, to improve the writing fluency of the paper. All scientific content, analyses, and conclusions were conceived, validated, and interpreted by the authors.

\section{More Visualization Results}
\label{appendix-samples}
\begin{figure*}[t]
    \centering
    \includegraphics[width=\linewidth]{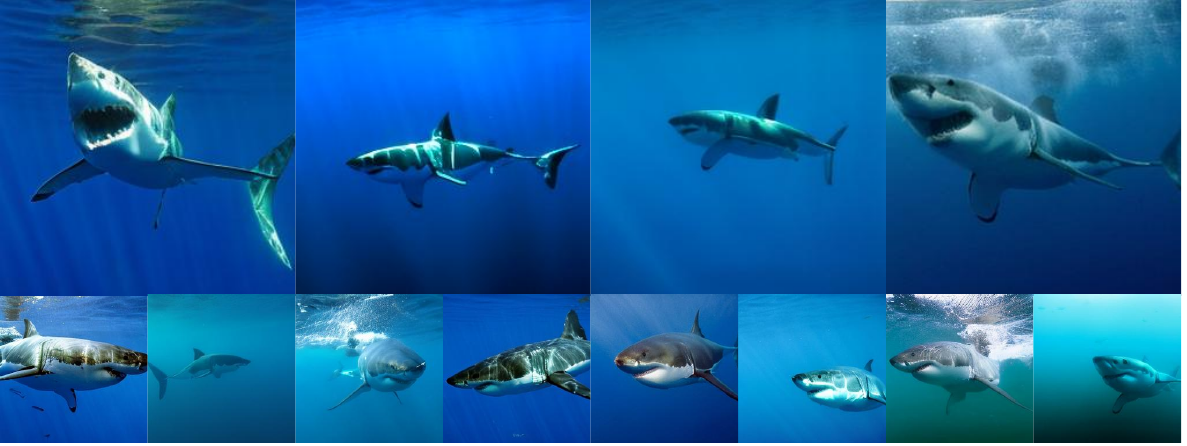} 
    \caption{The visualization results of SiT-XL/2 + sREPA utilize Classifier-Free Guidance (CFG) with $w = 4.0$ for the class ``Great white shark'' (index 2).}
    \label{Appendix-samples-2}
\end{figure*}

\begin{figure*}[t]
    \centering
    \includegraphics[width=\linewidth]{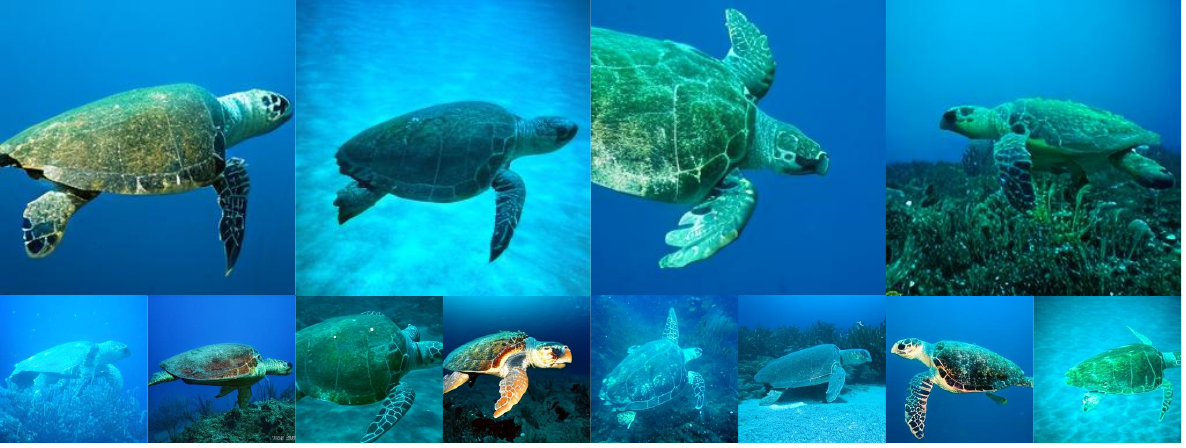} 
    \caption{The visualization results of SiT-XL/2 + sREPA utilize Classifier-Free Guidance (CFG) with $w = 4.0$ for the class ``loggerhead turtle'' (index 33).}
    \label{Appendix-samples-33}
\end{figure*}

\begin{figure*}[t]
    \centering
    \includegraphics[width=\linewidth]{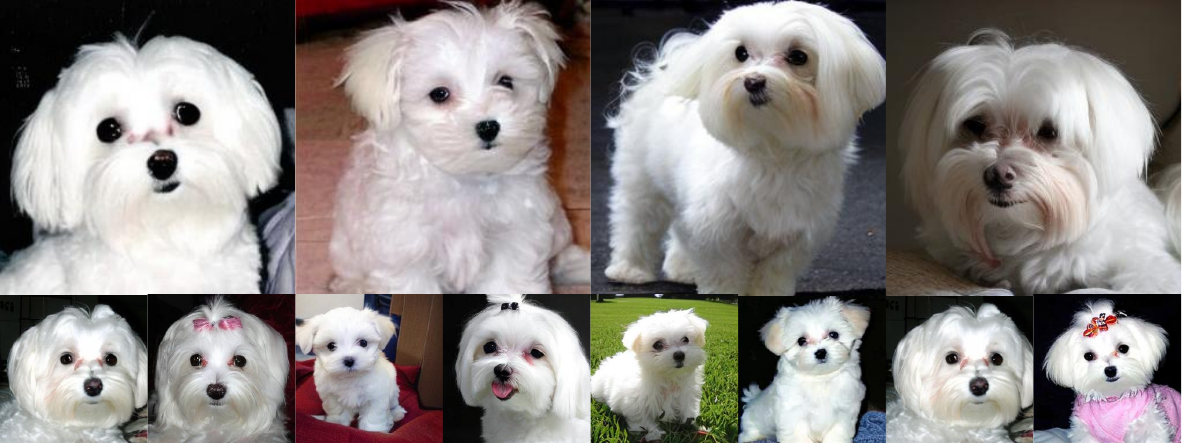} 
    \caption{The visualization results of SiT-XL/2 + sREPA utilize Classifier-Free Guidance (CFG) with $w = 4.0$ for the class ``Maltese dog'' (index 153).}
    \label{Appendix-samples-153}
\end{figure*}

\begin{figure*}[t]
    \centering
    \includegraphics[width=\linewidth]{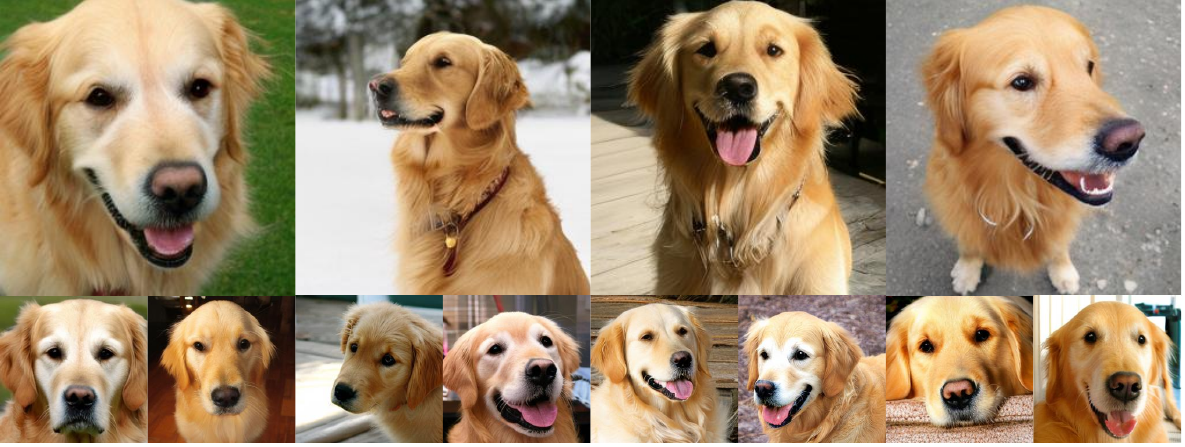} 
    \caption{The visualization results of SiT-XL/2 + sREPA utilize Classifier-Free Guidance (CFG) with $w = 4.0$ for the class ``golden retriever'' (index 207).}
    \label{Appendix-samples-207}
\end{figure*}

\begin{figure*}[t]
    \centering
    \includegraphics[width=\linewidth]{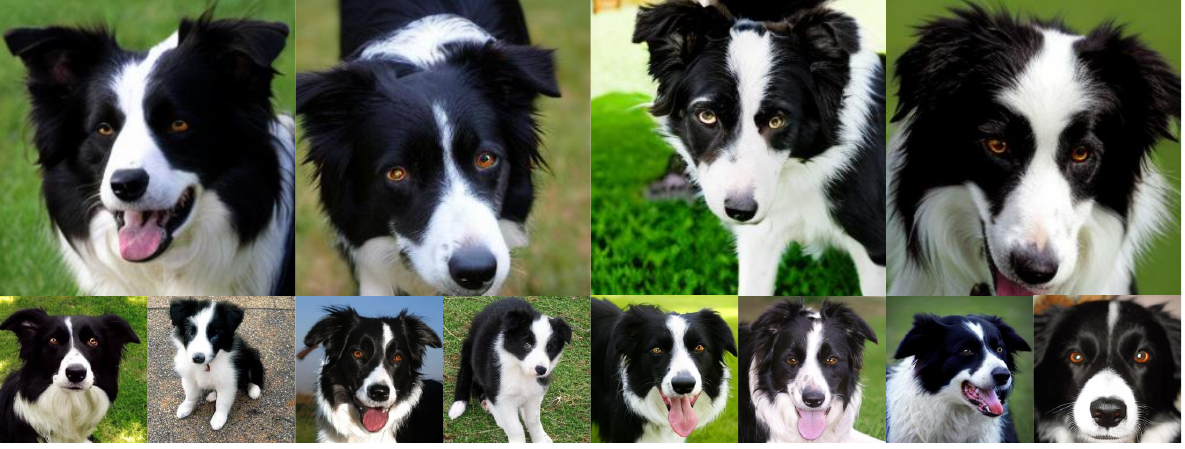} 
    \caption{The visualization results of SiT-XL/2 + sREPA utilize Classifier-Free Guidance (CFG) with $w = 4.0$ for the class ``Border collie'' (index 232).}
    \label{Appendix-samples-232}
\end{figure*}

\begin{figure*}[t]
    \centering
    \includegraphics[width=\linewidth]{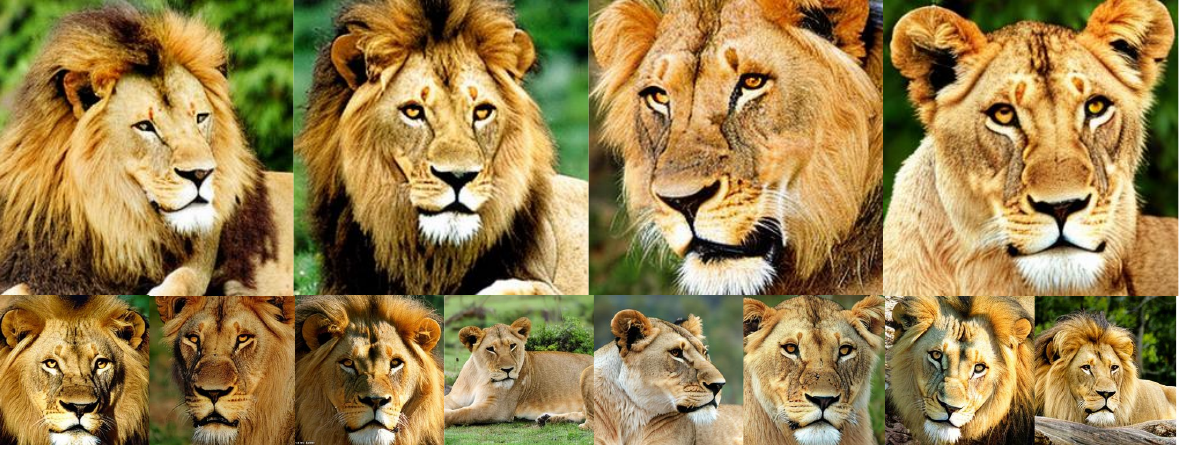} 
    \caption{The visualization results of SiT-XL/2 + sREPA utilize Classifier-Free Guidance (CFG) with $w = 4.0$ for the class ``lion'' (index 291).}
    \label{Appendix-samples-291}
\end{figure*}

\begin{figure*}[t]
    \centering
    \includegraphics[width=\linewidth]{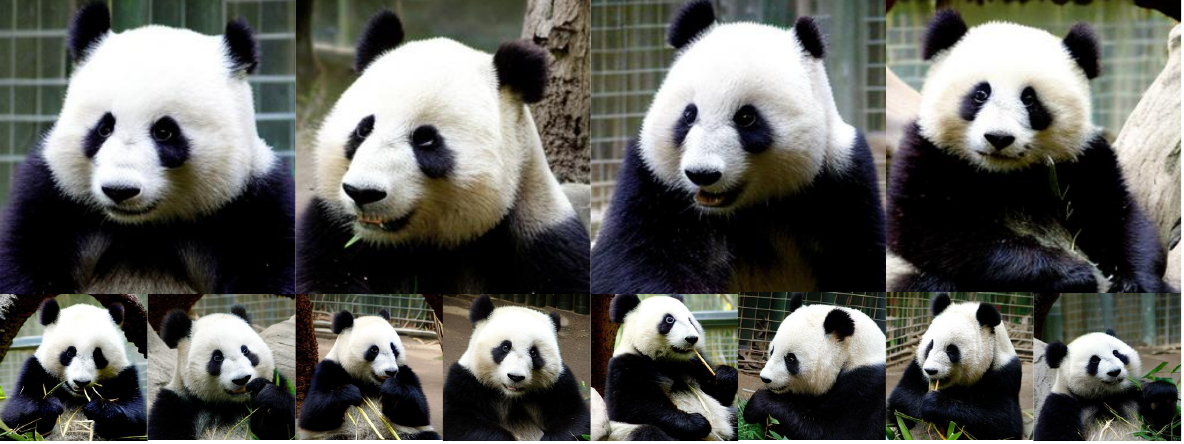} 
    \caption{The visualization results of SiT-XL/2 + sREPA utilize Classifier-Free Guidance (CFG) with $w = 4.0$ for the class ``panda'' (index 388).}
    \label{Appendix-samples-388}
\end{figure*}

\begin{figure*}[t]
    \centering
    \includegraphics[width=\linewidth]{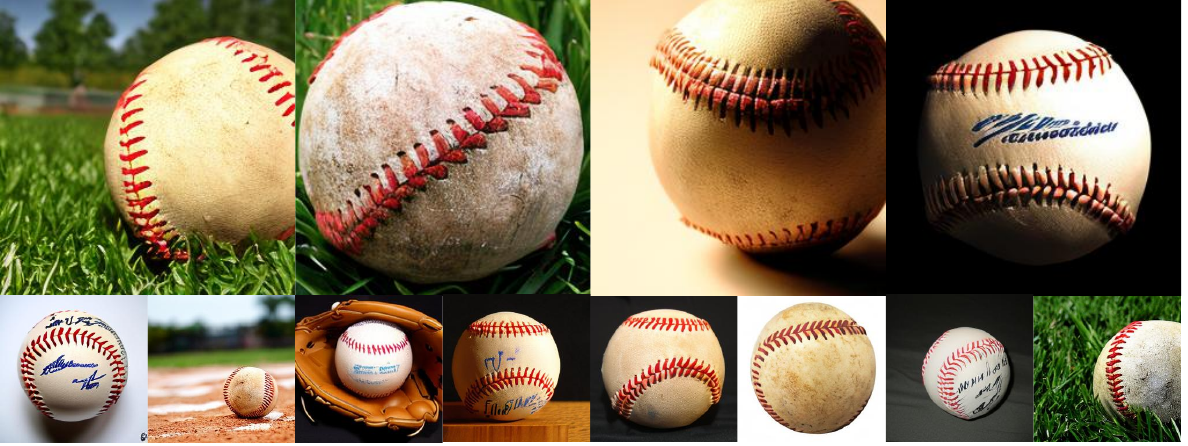} 
    \caption{The visualization results of SiT-XL/2 + sREPA utilize Classifier-Free Guidance (CFG) with $w = 4.0$ for the class ``baseball'' (index 429).}
    \label{Appendix-samples-429}
\end{figure*}

\begin{figure*}[t]
    \centering
    \includegraphics[width=\linewidth]{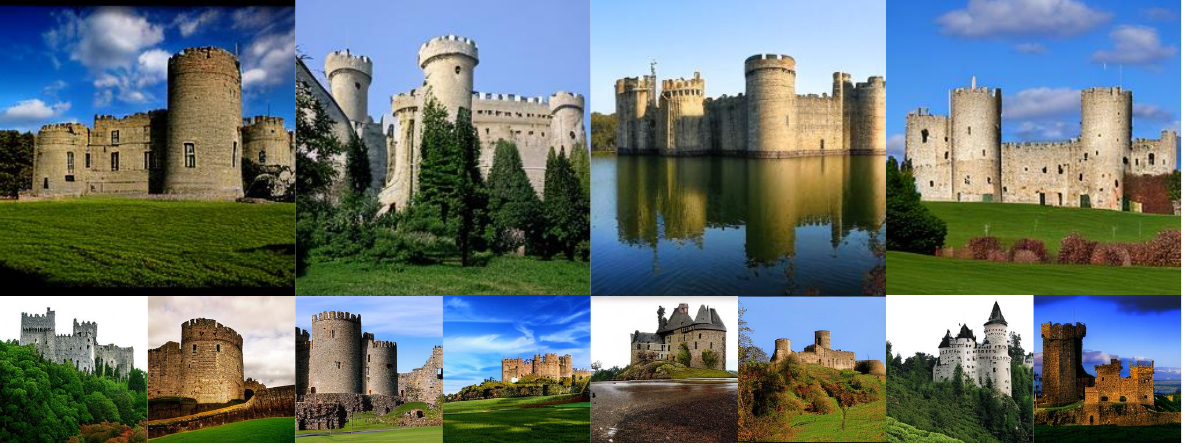} 
    \caption{The visualization results of SiT-XL/2 + sREPA utilize Classifier-Free Guidance (CFG) with $w = 4.0$ for the class ``castle'' (index 483).}
    \label{Appendix-samples-483}
\end{figure*}

\begin{figure*}[t]
    \centering
    \includegraphics[width=\linewidth]{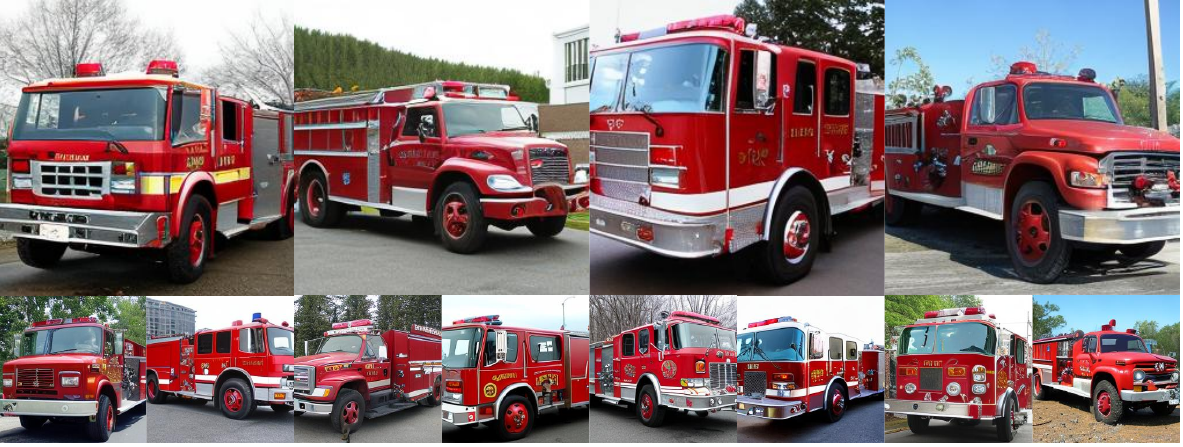} 
    \caption{The visualization results of SiT-XL/2 + sREPA utilize Classifier-Free Guidance (CFG) with $w = 4.0$ for the class ``fire truck'' (index 555).}
    \label{Appendix-samples-555}
\end{figure*}

\begin{figure*}[t]
    \centering
    \includegraphics[width=\linewidth]{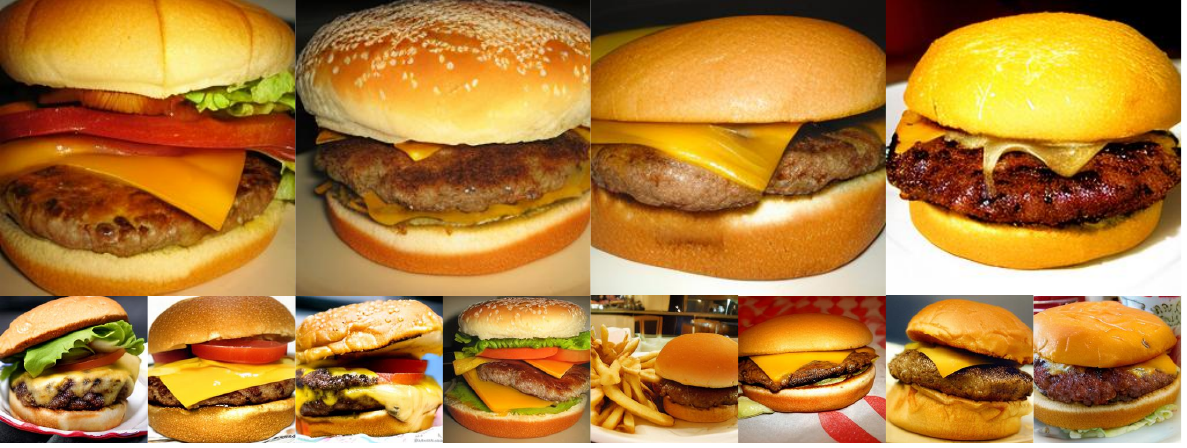} 
    \caption{The visualization results of SiT-XL/2 + sREPA utilize Classifier-Free Guidance (CFG) with $w = 4.0$ for the class ``cheeseburger'' (index 933).}
    \label{Appendix-samples-933}
\end{figure*}

\begin{figure*}[t]
    \centering
    \includegraphics[width=\linewidth]{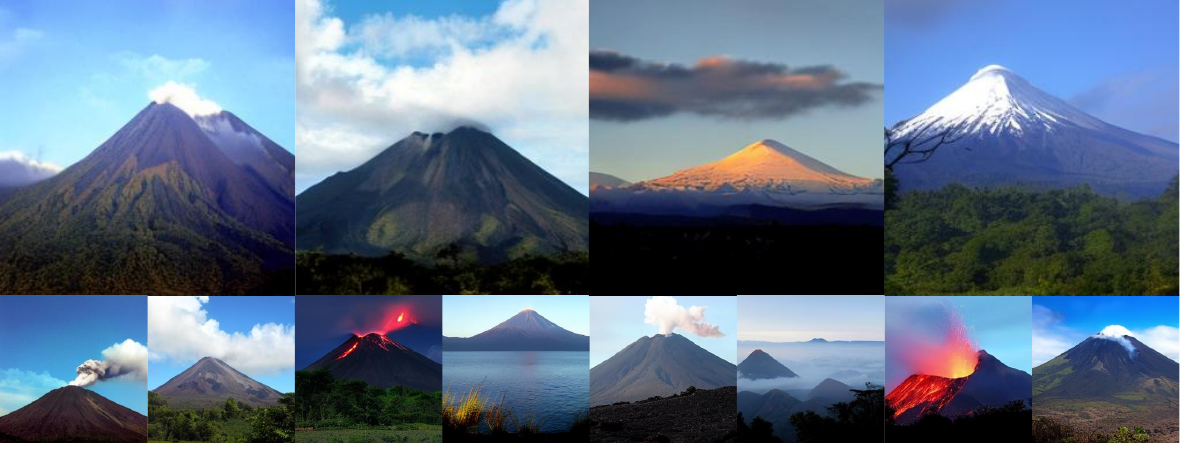} 
    \caption{The visualization results of SiT-XL/2 + sREPA utilize Classifier-Free Guidance (CFG) with $w = 4.0$ for the class ``volcano'' (index 980).}
    \label{Appendix-samples-980}
\end{figure*}

We present more visualization results of sREPA in Figure ~\ref{Appendix-samples-2}-\ref{Appendix-samples-980} with CFG $w = 4.0$.

\end{document}